\documentclass[conference]{IEEEtran}
\IEEEoverridecommandlockouts
\usepackage{cite}
\usepackage{amsmath,amssymb,amsfonts}
\usepackage{algorithm}
\usepackage{algcompatible}

\usepackage{algorithmicx}

\usepackage{wrapfig}

\usepackage{graphicx}
\usepackage{enumitem}
\usepackage{xcolor}
\usepackage{color}
\usepackage{url}
\usepackage{atbegshi}
 \def\BibTeX{{\rm B\kern-.05em{\sc i\kern-.025em b}\kern-.08em T\kern-.1667em\lower.7ex\hbox{E}\kern-.125emX}}

\renewcommand{\baselinestretch}{.95}

\usepackage{multicol}
\usepackage{verbatim}

\makeatletter
\newcommand\fs@spaceruled{\def\@fs@cfont{\bfseries}\let\@fs@capt\floatc@ruled
 \def\@fs@pre{\vspace{5\baselineskip}\hrule height.8pt depth0pt \kern2pt}%
 \def\@fs@post{\kern2pt\hrule\relax}%
 \def\@fs@mid{\kern2pt\hrule\kern2pt}%
 \let\@fs@iftopcapt\iftrue}
\makeatother

\begin{document}

\title{Particle Swarm Optimization Driven Federated Learning: Smart City and IIoT Case Studies\\}
\title{Particle Swarm Optimized Federated Learning:\\ Smart City and Industrial IoT Case Studies}
\title{Particle Swarm Optimized Federated Learning For Industrial IoT and Smart City Services}

\author{\IEEEauthorblockN{Basheer Qolomany}
\IEEEauthorblockA{\textit{Department of Cyber Systems} \\
\textit{University of Nebraska at Kearney}\\
Kearney, NE, USA \\
qolomanyb@unk.edu}
\and
\IEEEauthorblockN{Kashif Ahmad}
\IEEEauthorblockA{\textit{ Information \& Computing Technology Division } \\
\textit{Hamad Bin Khalifa University (HBKU)}\\
Doha, Qatar \\
kahmad@hbku.edu.qa}
\and
\IEEEauthorblockN{Ala Al-Fuqaha}
\IEEEauthorblockA{\textit{ Information \& Computing Technology Division} \\
\textit{Hamad Bin Khalifa University (HBKU)}\\
Doha, Qatar \\
aalfuqaha@hbku.edu.qa}
\and
\IEEEauthorblockN{Junaid Qadir}
\IEEEauthorblockA{\textit{Information Technology University} \\
Lahore, Pakistan \\
junaid.qadir@itu.edu.pk}
}

\author{\IEEEauthorblockN{Basheer Qolomany\IEEEauthorrefmark{1},
Kashif Ahmad\IEEEauthorrefmark{2},
Ala Al-Fuqaha\IEEEauthorrefmark{2}, and
Junaid Qadir\IEEEauthorrefmark{3}}
\IEEEauthorblockA{\IEEEauthorrefmark{1}Department of Cyber Systems, College of Business \& Technology,
University of Nebraska at Kearney, \\ Kearney, NE 68849 USA, qolomanyb@unk.edu}
\IEEEauthorblockA{\IEEEauthorrefmark{2}Information \& Computing Technology Division, College of Science and Engineering, Hamad Bin Khalifa University, \\ Doha, Qatar, \{kahmad, aalfuqaha\}@hbku.edu.qa}
\IEEEauthorblockA{\IEEEauthorrefmark{4}Department of Electrical Engineering, Information Technology University, Lahore, Pakistan, junaid.qadir@itu.edu.pk}}

\maketitle

\begin{abstract}
Most of the research on Federated Learning (FL) has focused on analyzing global optimization, privacy, and communication, with limited attention focusing on analyzing the critical matter of performing efficient local training and inference at the edge devices. One of the main challenges for successful and efficient training and inference on edge devices is the careful selection of parameters to build local Machine Learning (ML) models. To this aim, we propose a Particle Swarm Optimization (PSO)-based technique to optimize the hyperparameter settings for the local ML models in an FL environment. We evaluate the performance of our proposed technique using two case studies. First, we consider smart city services, and use an experimental transportation dataset for traffic prediction as a proxy for this setting. Second, we consider Industrial IoT (IIoT) services, and use the real-time telemetry dataset to predict the probability that a machine will fail shortly due to component failures. Our experiments indicate that PSO provides an efficient approach for tuning the hyperparameters of deep Long short-term memory (LSTM) models when compared to the grid search method. Our experiments illustrate that the number of clients-server communication rounds to explore the landscape of configurations to find the near-optimal parameters are greatly reduced (roughly by two orders of magnitude needing only 2\%-- 4\% of the rounds compared to state of the art non-PSO-based approaches).
We also demonstrate that utilizing the proposed PSO-based technique to find the near-optimal configurations for FL and centralized learning models does not adversely affect the accuracy of the models. 

\end{abstract}

\begin{IEEEkeywords}
Federated Learning, Deep Learning, Particle Swarm Optimization, Parameter Optimization, IoT, Smart Services
\end{IEEEkeywords}

\section{Introduction}


In recent years, the enormous growth in the processing and storage capabilities of end devices, and an ever-increasing concern about data privacy has led to a growing interest in Federated Learning (FL) \cite{yang2019federated}. FL, which builds a Machine Learning (ML) model across multiple decentralized edge devices, empowers multiple stakeholders (i.e., data owners) to benefit from each others' data by training a common and robust ML model collaboratively without exposing their data to other participants. The core concept of FL is based on training local models on the local resources (i.e., training samples) and then sharing the model's parameters with the other parties via a central server or in a peer-to-peer (P2P) manner depending on the FL topology to build a global model. Combining edge computing characteristics with data privacy, FL becomes an attractive proposition for a diverse set of applications---such as sentiment analysis, monitoring activities of mobile users, autonomous vehicles, and healthcare \cite{anguita2013public,huang2018loadaboost}, where data is distributed at multiple devices. 

The feasibility of FL is highly dependent on the capabilities of edge devices to perform efficient local training and inference with the performance also depending on the appropiate tuning of the ML models and their parameters. For instance, the performance of deep neural networks (DNNs) depends on parameters, such as the number of hidden layers, the number of neurons in a layer, and the total number of epochs \cite{qolomany2017parameters}. One of the main challenges in efficient training at the edge devices is the selection of the appropriate structure of the local ML algorithm and parameter settings.



The data privacy aspects and building a global model from distributed and heterogeneous data sources pose several challenges for FL in terms of optimization, data privacy, and efficient communication. In particular, existing works on FL assume fixed settings (i.e., architectures/structure and other parameters) for the ML models trained on the local dataset at the local hosts without exploring the optimal values for these parameters. Though there are several heuristic rules of thumb \cite{swingler1996applying,qolomany2017parameters}, there is no theory yet to define the structure of ML models for a particular application. The rules of thumb, such as ``\textit{number of hidden layers in NNs should always be less than or equal to the number of inputs to the NNs} \cite{swingler1996applying}'' and ``\textit{the number of hidden layer neurons should be less than twice of the number of neurons in the input layer} \cite{berry2004data}'' do not always guarantee better results as they do not consider different factors, such as the number of training epochs, noise in the data, and the complexity of the objective function \cite{qolomany2017parameters}. Moreover, the manual selection of these parameters using a trial and error method is cumbersome and requires an extensive experimental setup. 

 In this paper, we propose a Particle Swarm Optimization (PSO)-based parameter setting framework for the common parameters of the local models trained by edge devices on the local data in an FL environment. More specifically, we use PSO to optimize three different parameters of local models---namely, (i) the number of hidden layers ($L$), (ii) the number of neurons in each layer ($N$), and (iii) the number of epochs ($E$) representing the number of training passes each client makes over its local dataset in each round of the FL algorithm. 

PSO and other heuristic techniques introduce randomness and depend on luck to approach an optimal solution. They are designed to solve a problem within a reasonable computation time and produce a solution that is good enough for solving the problem at hand. Since there is no guarantee that the solution found using heuristic techniques will be the optimal solution, the solution is called ``sub-optimal.'' Our choice of PSO as an optimization method is motivated by its simplicity, easy implementation, robustness, and fast convergence. Such characteristics make it a preferred choice in diverse application domains \cite{ahmad2018ensemble,ahmad2019deep}. This work is inspired by our previous work \cite{qolomany2017parameters}, where we proposed a PSO-based optimization framework for parameter tuning of deep learning models. In this work, we use a similar concept to optimize a FL framework by exploring the potential solution space instead of assuming a fixed structure and parameters for the local ML models. 

The main contributions of the work can be summarized as:
\begin{itemize}
 \item [(i)] We propose a PSO-driving federated learning framework, where PSO-based optimization has been employed to tune the common parameters of the local models trained on edge devices.
 \item [(ii)] We also evaluate and compare the proposed PSO-based parameter optimization approach with the grid search technique to find the best configurations for local models in an FL environment.
 \item [(iii)] We also analyze the error rates of the best configurations found using the proposed PSO and grid search for FL and centralized learning models. 
\end{itemize}

\section{Related Work}
\label{sec:related_work}



Several interesting optimization techniques have been proposed to deal with the challenges associated with learning a ML model in FL environment (such as the non-IID, unbalanced, and highly distributed nature of data) \cite{mcmahan2017communication,li2019federated,sahu2018federated}. For instance, to deal with unbalanced and non-IID data, Mcmahan et al. \cite{mcmahan2017communication} proposed a synchronous update scheme called FedAvg for the optimization of the global model, where a fraction of clients are selected among the fixed number of clients each with a fixed local dataset by the sever to share the current global model's parameters. The selected clients then perform computation based on global parameters and their local dataset and resend the updates to the server. The server then uses these updates to compute the global model by averaging local Stochastic Gradient Descent (SGD) updates and repeats the process. Though the FedAvg technique has proved itself to be effective in dealing with unbalanced and non-IID data, it does not guarantee convergence in the presence of heterogeneous data \cite{sahu2018federated}. To deal with heterogeneous data in FL, Sahu et al. \cite{sahu2018federated} proposed a modified version of FedAvg, namely FedProx, by adding a proximal term to the objective function of FedAvg, which helps to improve the stability of the method against heterogeneous data. On the other hand, in \cite{smith2017federated}, a multi-task learning-based solution, namely MOCHA, has been proposed to deal with the statistical challenges associated with FL by learning separate but related models for each node, simultaneously. Several other interesting solutions have been proposed to ensure convergence in the presence of heterogeneous data using assumptions such as convexity and uniformly bounded gradients \cite{wang2019adaptive,li2019federated}.

There is a lot of research literature dealing with parameter optimization of ML models with different techniques being proposed. For instance, Tran et al. \cite{tran2020hyper} analyzed the impact of parameters optimization on the performance of traditional classification algorithms. In this regard, several optimization techniques have also been employed for optimizing the parameters of Support Vectors Machines (SVMs) \cite{wang2020parameters,tharwat2019quantum,rajalaxmi2019mutated}. 
Several methods have also been proposed for the optimization of different parameters of Artificial Neural Networks (ANNs) \cite{hassim2013solving,zhou2020employing,ccam2015learning}. Some of the methods aim at optimizing the initial weights of NNs. For instance, in \cite{hassim2013solving}, instead of the traditional Back-propagation (BP) strategy for Functional Link Neural Networks (FLNN), a Bee Colony (BC) optimization algorithm has been employed to set the weights of the NNs. BC along with PSO-based methods have also been employed in \cite{zhou2020employing}, where both methods are used independently and in combination to optimize the computational parameters (i.e., weights and biases) for a Multi-Layer Perceptron (MLP) neural network. Apart from the initial weights and biases, the performance and the feasibility of NNs in an application also depends on other parameters, such as the number of hidden layers, learning rate, and momentum coefficients. To this aim, several techniques have also been proposed to optimize these parameters. For instance, Mirjalili et al. \cite{mirjalili2012training} proposed a hybrid PSO and gravitational search algorithm to optimize the learning rate and momentum of NNs. Similarly, Francesco et al. \cite{junior2019particle} proposed psoCNN, an optimized CNN architecture for image classification where PSO is used to identify an optimized CNN architecture achieving better performance against state-of-the-art architectures. In \cite{serizawa2020optimization}, a linearly decreasing weight particle swarm optimization (LDWPSO) method is proposed for optimizing hyperparameters of a CNN architecture.

The literature depicts that most of the existing works focus on global optimization, privacy, and communication aspects to analyze the feasibility of FL. However, we believe that the feasibility of FL is also highly constrained by the capabilities of edge devices to perform efficient local training and inference. Despite its importance, no prior work has analyzed this aspect of FL. To the best of our knowledge, the paper is the first work to focus on this important yet overlooked aspect of FL.

\section{PSO-Based Parameter Optimization Model}
\label{sec:PSO}

The PSO was originally developed in 1995 by Kennedy and Eberhart \cite{kennedy_particle_1995}, and inspired by the behavior of animal groups, such as bird and fish swarms \cite{clerc_particle_2010}. The PSO is an iterative optimization method that has extensive capabilities for global optimization in addition to easy implementation, scalability, robustness, and fast convergence. It employs simple mathematical operators and is computationally inexpensive in terms of both memory requirements and speed \cite{qolomany2017parameters}. The PSO algorithm is composed of a group of particles called a swarm, each particle represents a possible solution to the problem. These particles in the swarm repeatedly communicate with each other. Each particle is associated with its position, velocity and fitness value that is determined by an optimization function. The position of each particle in the swarm is updated to move closer to the particle which has the best position. While the velocity value conveys the direction and distance of the particle's movement.

In the proposed work, the process starts by initializing a group of random particles for $L$, $N$, and $E$, then it searches for an optima by updating generations. In every iteration, each particle is updated by following the two ``best'' values. The first one is the best solution (accuracy) it has achieved so far. This value is called \textit{pbest} (the particle best position). Another best value that is tracked by the particle swarm optimizer is the best value obtained so far by any particle in the population. This best value is a global best and called \textit{gbest} (the global best position). Each particle updates its velocity and position by tracking the values of \textit{pbest} and \textit{gbest} in each iteration. 

The velocity and position of each particle for the number of hidden layers are updated according to the Equations (1) and (2):
\begin{equation}
\begin{split}
V_{L,i}^{t+1}= w. V_{L,i}^t + c_1.rand.(L_{i}^{best} - V_{L,i}^t) \\+ c_2.rand.(G^{Lbest}-V_{L,i}^t)
\end{split}
\end{equation}
Where $V_L$ is the velocity of the number of hidden layers, $L_i^{best}$ is the particle's best local value of the number of hidden layers, and $G^{Lbest}$ is the best global value of the number of hidden layers. The position for the number of hidden layers of particle $i$ is updated based on the particle's velocity $V_L$ in Eq(2):
\begin{equation}
L_i^{t+1}= L_i^t + V_{L,i}^{t+1}
\end{equation}
The velocity and position of each particle for the number of neurons in each layer layer are updated according to the Equations (3) and (4) as follows:
\begin{equation}
\begin{split}
V_{N,i}^{t+1}= w. V_{N,i}^t + c_1.rand.(N_{i}^{best} - V_{N,i}^t) \\+ c_2.rand.(G^{Nbest}-V_{N,i}^t)
\end{split}
\end{equation}
Where $V_N$ is the velocity of the number of neurons in each hidden layer, $N_i^{best}$ is the particle's best local value of the number of neurons in each hidden layer, and $G^{Nbest}$ is the best global value of the number of neurons in each hidden layer. The position for the number of neurons of particle $i$ is updated based on the particle's velocity $V_N$ in Eq. (4):
\begin{equation}
N_i^{t+1}= N_i^t + V_{N,i}^{t+1}
\end{equation}
The velocity and position of each particle for the number of epochs are updated according to the Equations (5) and (6) as follows:
\begin{equation}
\begin{split}
V_{E,i}^{t+1}= w. V_{E,i}^t + c_1.rand.(E_{i}^{best} - V_{E,i}^t) \\+ c_2.rand.(G^{Ebest}-V_{E,i}^t)
\end{split}
\end{equation}
Where $V_E$ is the velocity of the number of epochs, $E_i^{best}$ is the particle's best local value of the number of epochs, and $G^{Ebest}$ is the best global value of the number of epochs. The position for the number of epochs of particle $i$ is updated based on the particle's velocity $V_E$ in Eq. (6):
\begin{equation}
E_i^{t+1}= E_i^t + V_{E,i}^{t+1}
\end{equation}
In Equations (1), (3) and (5), $c_1$ and $c_2$ are the acceleration coefficients and represent the weights of approaching the $pbest$ and $gbest$ of a particle. $w$ is the inertia coefficient as it helps the particles to move by inertia towards better positions. $rand$ is a uniform random value between $0$ and $1$. The parameters utilized in our experiments are listed in Table I.

\begin{table}
\centering
\caption{THE PARAMETERS UTILIZED IN OUR EXPERIMENTS}
\label{Table_1}
 \begin{tabular}{|p{3.67cm}|p{4.55cm}|} 
 \hline
 Parameter & Value \\ [0.5ex] 
 \hline\hline
 Population size & 5 \\
 \hline
 Learning coefficient, $c_1, c_2$ & uniformly distributed between [0,4] \\
 \hline
 Maximum number of iterations & 10 \\
 \hline
 Number of hidden layers & within the range [1, 5] increment by 1  \\
 \hline
 Number of neurons in each layer& within the range [1, 200] step size varied \\
 \hline
 Number of epochs & within the range [1, 50] increment by 5\\
 \hline
 Number of clients & 5\\
 \hline
Number of clients-server communication rounds (CommRounds) for a configuration settings &15\\
\hline
 Total Number of clients-server communication rounds & = CommRounds $\times$ the number of tried model configurations to get the best\\
 
 \hline
Hidden layer velocity & $\bullet$$MinLayerVelocity= \hfill -0.1 \times (MaxLayers - MinLayers)$

$\bullet$$MaxLayerVelocity= +0.1 \times (MaxLayers - MinLayers) $ \\
 \hline
 Neuron velocity & $\bullet$$MinLayerVelocity= -0.1 \times (MaxNeurons - MinNeurons)$

$\bullet$$MaxLayerVelocity= +0.1 \times (MaxNeurons - MinNeurons) $ \\
 \hline
 Epoch velocity & $\bullet$$MinLayerVelocity= -0.1 \times (MaxEpochs - MinEpochs)$

$\bullet$$MaxLayerVelocity= +0.1 \times (MaxEpochs - MinEpochs) $ \\[1ex]
 \hline
\end{tabular}
\end{table}

Algorithm \ref{alg_1_PSO} describes our proposed PSO-based parameter selection technique for deep LSTM models. The server starts by initializing random parameter configurations for the number of hidden layers, the number of neurons in each layer, and the number of epochs of deep LSTM models. The algorithm returns the near-optimal configuration. The algorithm is presented for two use-case scenarios (i.e., smart city services and the IIoT services proxy use cases), which will be fully explored in Section \ref{sec:experiments}.

\begin{algorithm}[t]
\caption{PSO for Parameter Optimization of Deep LSTM Models. }

\label{alg_1_PSO}
\begin{algorithmic}[1]
\STATEx \textbf{Input}: A dataset. Values of acceleration constants ($c_1$ and $c_2$); $w; MaxIt$ (the maximum number of iterations to reach a good solution); and the range bounds $MinLayer$, $MaxLayer$, $MinNeurons$, $MaxNeurons$, $MaxLayerVelocity$ and $MaxNeuronVelocity$ (set as described in Table \ref{Table_1}). 
\STATEx \textbf{Output}: Near-optimal configuration in terms of $L$, $N$ and $E$ for the deep LSTM model.
\STATEx $\qquad \qquad \qquad \qquad \qquad$
\STATEx \textbf{Begin}
\STATE Initialization
\begin{enumerate}[label=\alph*.]
\item Initialize $L_i$, $N_i$, and $E_i$ positions randomly for the number of layers, number of neurons, and number of epochs, respectively. 
\item Initialize $V_L$, $V_N$, and $V_E$ velocities randomly for the number of layers, number of neurons, and number of epochs, respectively. 
\item Define the fitness function (model accuracy for classification problem or root mean square error (RMSE) for the regression problem). 
\item 	Calculate the fitness value for each particle and set $pbest$ for each particle and $gbest$ for the population. 
\end{enumerate}
\STATE Repeat the following steps until the $gbest$ solution does not change anymore or the $MaxIt$ is reached.
\begin{enumerate}[label=\alph*.]
\item Update $L_i$, $N_i$, $E_i$, $V_L$, $V_N$, and $V_E$ in each particle according to the Equations (1) through (6). 
\item Calculate the fitness value for each particle. If the fitness value of the new location is better than $pbest$, the new location is updated to be the $pbest$ location.
\item If the currently best particle in the population is better (e.g. in terms of Accuracy/RMSE) than the $gbest$, the best particle replaces the recorded $gbest$.
\end{enumerate}
\STATE Return the near-optimal $L$, $N$ and $E$ for the LSTM model. 
\STATEx \textbf{End}
\end{algorithmic} 
\end{algorithm}

\section{Datasets}
\label{sec:dataset}
In the following subsections, we describe the datasets used for the experimental evaluations of the proposed work along with the details of the preprocessing of the data.

\subsection{Smart City Traffic Flow Prediction Case Study}

Traffic flow prediction plays an important role in intelligent transportation management and route guidance. By collecting and analyzing the traffic information, many benefits can be brought from traffic predictions, for instance, it can help in relieving traffic congestion, reducing air pollution, providing road safety and secure traffic conditions \cite{lv_traffic_2015} \cite{qolomany_leveraging_2019}. However, since a vehicle's location is tightly bundled with its driver, it may threaten vehicles' location and trajectory privacy. An attacker can predict a driver's future location based on his vehicle's trajectory, or even infer the drivers' personal information, such as habits, health condition, income, and religious belief, according to their frequently visited places. 

The aforementioned aspects of the task motivated us to deploy our proposed PSO-driven FL framework to the task where the City Pulse EU FP7 project dataset \cite{noauthor_citypulse_nodate} is used for traffic prediction. This dataset conveys the vehicular traffic volume collected from the city of Aarhus, Denmark, observed between two points for a set duration of time of 6 months. The dataset is divided into training and testing subsets. The training set is sampled into 18 different non-overlapped samples. Each sample is assigned to a client, which uses it to locally train a deep LSTM model \cite{qolomany_role_2017}. This means that the local training dataset of a client is never uploaded to the server. Instead, all the clients are coordinated by a central server, such that each client computes an update to the current global model maintained by the server, and only this update is communicated.

\subsection{IIoT Predictive Maintenance Case Study}

A predictive maintenance (PM) strategy uses ML methods to identify, monitor, and analyze system variables during an operation. Also, PM alerts operators to preemptively perform maintenance before a system failure occurs \cite{sipos_log-based_2014} \cite{qolomany_trust-based_2020}. Being able to stay ahead of equipment shutdowns in a mine, steel mill, or factory, PM can reduce downtimes and maintenance costs for a busy enterprise. Such sensitive data leakage may cause a variety of cyber-attacks to Industrial IoT (IIoT) systems. One of the first successful attacks against industrial control systems was the Slammer worm, which infected two critical monitoring systems of a nuclear power plant in the U.S.A \cite{sadeghi_security_2015}. 

Our second case study is a proxy for IIoT  services in which the real-time telemetry data, provided by Microsoft Azure Intelligence Gallery \cite{noauthor_azure_nodate}, created by data simulation methods collected from 100 machines in real time averaged over every hour collected during the year 2015 is used. The data consists of voltage, rotation, pressure, vibration measurements, error messages, historical maintenance records that include failures, and machine information such as type and age.  The goal is to analyze predictive maintenance use-case problems caused by component failures such that the question ``What is the probability that a machine will fail in the near future due to a failure of a certain component?'' can be answered. Telemetry data comes with time-stamps that makes it suitable for calculating lagging features. In our experiments, 24 hours lag features are calculated. The prediction problem for our experiments is to estimate the probability that a machine will fail in the next 24 hours due to a certain component failure (component 1, 2, 3, or 4). All records within a 24 hours window before a failure of component 1 have $failure=comp1$, and so on for components 2, 3, and 4; all records not within 24 hours of a component failure have $failure=none$. The dataset is divided into training and testing subsets. The training dataset is sampled into $99$ different non-overlapped samples (each sample represents the dataset for one machine) each sample is then assigned to a client that is used to train a deep LSTM model locally. All the clients are coordinated by a central server by updating the current global model maintained by the server. 

\begin{figure*}[t!]
\centering
\includegraphics[width=1\textwidth, height=5.1cm]{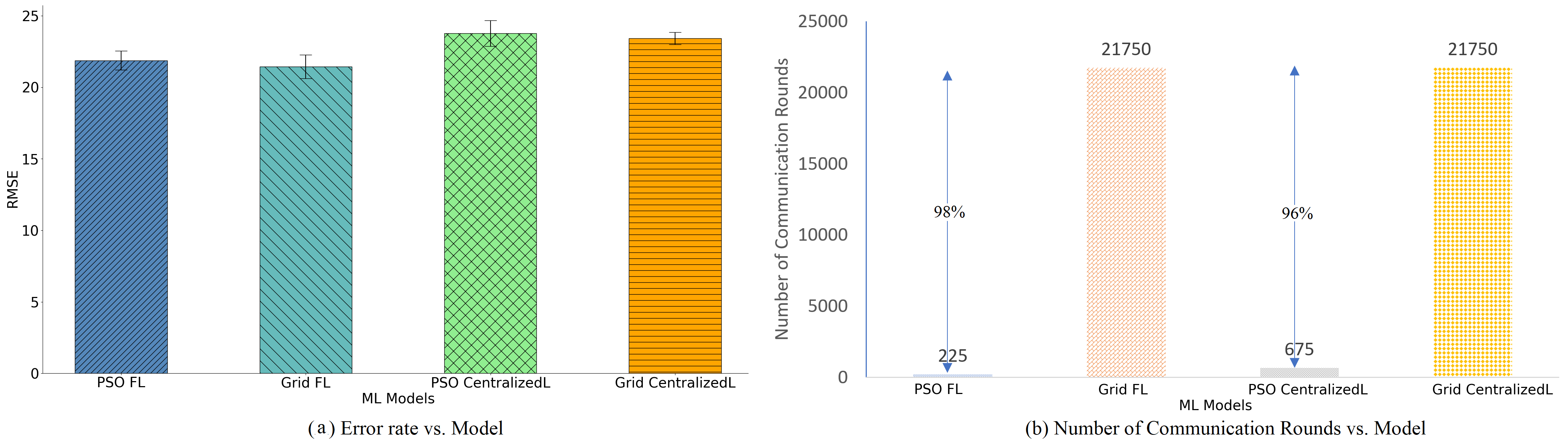}
\caption{Comparison between PSO and grid search in terms of \textbf{error rate} \& \textbf{number of communication rounds} for \textit{\textbf{smart city traffic prediction use-case}}. The PSO-based technique reduces the number of clients-server communication rounds by 96--98\% without adversely affecting the error rate of the models}
\label{Fig_12.png}
\end{figure*} 

\begin{figure*}[t!]
\centering
\includegraphics[width=1\textwidth, height=5.1cm]{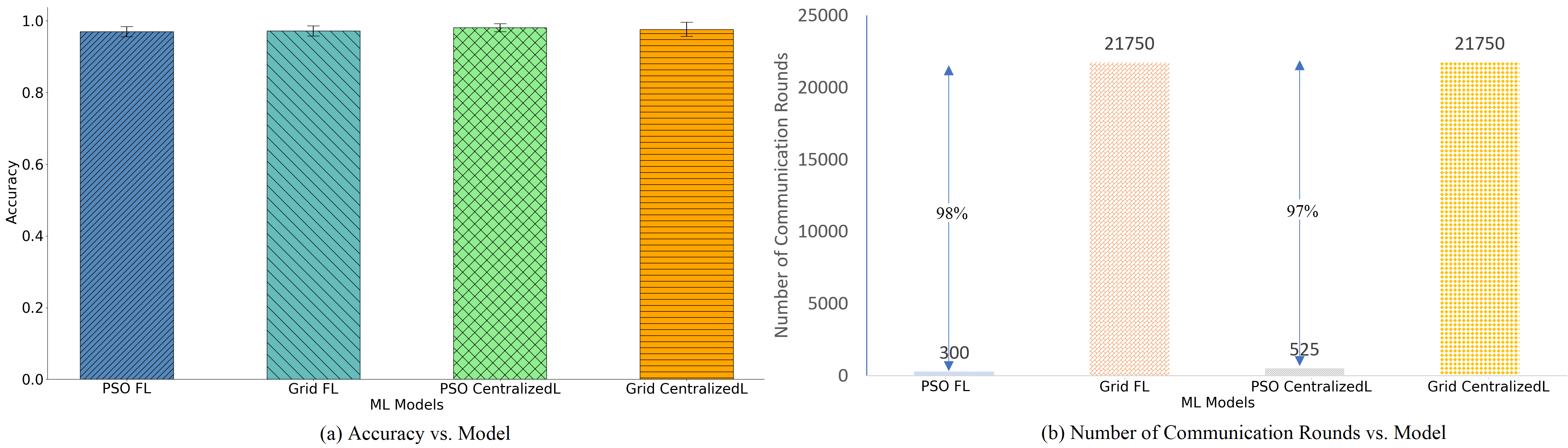}
\caption{Comparison between PSO and grid search in terms of \textbf{accuracy} \& \textbf{number of communication rounds} for \textit{\textbf{IIoT predictive maintenance use-case}}. The PSO-based technique reduces the number of clients-server communication rounds by 97--98\% without adversely affecting the accuracy of the models} 
\label{Fig_34.png}
\end{figure*} 

\section{Experimental Results}
\label{sec:experiments}

In our experiments, we selected two case studies to assess the performance of our proposed PSO-based parameter selection technique. We formulated the traffic prediction service as a regression problem, when LSTM is used to predict the number of cars, we use RMSE to measure LSTM model performance. While the predictive maintenance service as a classification problem to predict the probability that a machine will fail in the next $24$ hours due to component failure, we use accuracy to measure LSTM model performance. We then set out to address the main goal of this paper which is to compare our proposed PSO-based parameter selection technique vis-à-vis the grid search technique in terms of finding the best $L$, $N$, and $E$ for the LSTM models.
  
To evaluate and compare the grid search and PSO approaches, both the accuracy and the number of client-server communication rounds to explore the landscape of configurations to find the near-optimal parameters are considered. In the case of PSO, we consider the PSO technique with $5$ particles. The algorithm terminates when the maximum number of iterations is reached or when there is no difference between the accuracies of two consecutive iterations. In our experiments, we also compare the proposed PSO against a grid search-based approach to find the best configurations for the models in the FL environment versus centralized learning. Centralized learning refers to the scenario in which users' data is uploaded to a centralized server for training, and then re-deploying an iterative model back to the user. 

\subsection{Smart City Traffic Flow Prediction Results}
Figure \ref{Fig_12.png}-(a) shows the error rates of the best configurations using the proposed PSO and grid search in FL and centralized learning in traffic flow prediction use-case. This figure shows the RMSE as a 95\% confidence interval for the best five configurations obtained by PSO and grid search approaches. Figure \ref{Fig_12.png}-(b) shows the number of clients-server communication rounds that need to convergence the globally best solution for FL models in in traffic flow prediction use-case.

\subsection{IIoT Predictive Maintenance Results}

Figure \ref{Fig_34.png}-(a) shows the accuracy of the best configurations using the proposed PSO and grid search in FL and centralized learning schemes in predictive maintenance use-case. This figure shows the accuracy as a 95\% confidence interval for the best five configurations obtained by PSO and grid search approaches. 

Figure \ref{Fig_34.png}-(b) shows the number of clients-server communication rounds that need to convergence the globally best solution for FL models in IIoT predictive maintenance use-case.

\section{Findings and Lessons learned}
\label{sec:lessons_learned}

Based on our results, our findings and lessons learned are summarized next.

\begin{enumerate}

\item Our PSO-based technique can return impressive performance for parameter tuning of deep LSTM models and improve upon the results of grid search. 

\item Our proposed approach strives to decrease the exploration process of the landscape of configurations to find the near-optimal parameters without affecting the accuracy of the selected models, as Figures 1 and 2 indicate.

\item Figures \ref{Fig_12.png} and \ref{Fig_34.png} illustrate that the number of clients-server communication rounds can be massively decreased (to around $2\%$--$4\%$ of their original values returning a two-order of magnitude performance improvement) using our proposed PSO-based parameter value selection technique when compared with grid search method, while using PSO technique to find near-optimal configurations does not adversely affect the accuracy of the models trained in FL compared with centralizing learning approaches. 

\item Figures \ref{Fig_12.png}-(a) and \ref{Fig_34.png}-(a) show that using the proposed PSO technique is not affecting the accuracy of the models trained in federated and centralized learning. 

\item Figures \ref{Fig_12.png}-(b) and \ref{Fig_34.png}-(b) show that the number of clients-server communication rounds to explore the landscape of configurations to find the near-optimal parameters is decreased by around two orders of magnitude (to around $2\%$--$4\%$ of the rounds) using the PSO-based approach compared to the grid search method in both decentralized FL-based approach and centralized learning approach. 

\end{enumerate}

\section{Conclusion and Future Work}
\label{sec:conclusions}

Since the bottlenecks in the performance of the local models trained at the edge hinder the success of Federated Learning (FL), we have proposed in this paper a Particle Swarm Optimization (PSO)-based technique to optimize the hyperparameter settings for the deep Long Short-Term Memory (LSTM) models trained at the edge in an FL environment. The parameters we optimized in the work include the number of hidden layers, the number of neurons in each layer for deep LSTM, and the number of epochs which is the number of training passes each client makes over its local dataset on each round. The proposed method is evaluated in two use-cases. In the first case study, we consider traffic prediction models as a proxy of smart city services. While in the second case study, we consider predictive maintenance models as a proxy for industrial IoT (IIoT) services. The results obtained show that the number of client-server communication rounds to explore the landscape of configurations to find the near-optimal parameter settings is greatly decreased by two orders of magnitude using the PSO-based approach compared to the grid search method. Our results also show that using the proposed PSO technique to find the near-optimal configurations without affecting the accuracy of the models in FL and centralized learning. As a future extension, we intend to explore the use of PSO to tune other deep learning parameters (e.g., batch size, learning rate) as well as the other parameters that are associated with FL, such as the fraction of number of clients that the server communicates within each communication round and the number of federated learning communication rounds.


\bibliographystyle{IEEEtran}
\bibliography{References}

\end{document}